\documentclass{article}
\usepackage[utf8]{inputenc}
\usepackage[margin=1in]{geometry}
\usepackage{graphicx}
\usepackage{color} 
\usepackage{xcolor}
\usepackage{url}
\usepackage{hyperref}
\usepackage{amsthm}
\usepackage{amsmath}
\usepackage{amsfonts}
\usepackage{algorithm}
\usepackage{algpseudocode}
\usepackage[
backend=biber,
style=ieee,
sorting=ynt
]{biblatex}
\usepackage{stmaryrd}
\usepackage{imakeidx}
\theoremstyle{plain}

\theoremstyle{remark}

\newcommand{\R}{\mathbb{R}}

\newcommand{\vect}[1]{\mathbf{#1}}

\usepackage{authblk}
\algnewcommand\Yield{\textbf{yield }}
\makeindex
\title{\textbf{Measuring Classification Decision Certainty and Doubt}}

\author[1]{Alexander M. Berenbeim}
\author[1]{Iain J. Cruickshank}
\author[2]{Susmit Jha}
\author[1]{Robert H. Thomson}
\author[1]{Nathaniel D. Bastian}
\affil[1]{Army Cyber Institute, United States Military Academy, West Point, NY 10996, USA}
\affil[2]{Computer Science Laboratory, SRI International, Menlo Park, CA 94025, USA}

\date{}

\begin{document}
\maketitle
\begin{abstract} Quantitative characterizations and estimations of uncertainty are of fundamental importance in optimization and decision-making processes. Herein, we propose intuitive scores, which we call \textit{certainty} and \textit{doubt}, that can be used in both a Bayesian and frequentist framework to assess and compare the quality and uncertainty of predictions in (multi-)classification decision machine learning problems.
\end{abstract}

\section*{Introduction} 

The rising use of artificial intelligence and machine learning technologies to power  intelligent systems has led to a growing desire for automating, accelerating, and enhancing decision-making processes in safety-critical applications. These technologies offer decision-makers the ability to gain an information and decision-making advantage at the speed of machines. However, in safety-constrained decision-making, it is crucial to estimate and factor in the level of certainty and doubt associated with each classification decision \cite{MilcomIoBT22}. In such scenarios, even small chances of risky outcomes may have a significant impact on classification decisions, regardless of the most probable predicted outcome, and safety-critical applications need to be sensitive to such tail probabilities \cite{Cobb2021, vaicenavicius19a}.

Furthermore, in the (multi-class) classification decision setting when model-assigned probabilities are close to uniformly distributed, we have cause to doubt the model's prediction, even if it is accurate. Moreover, from a theoretical view of probability, we have cause to doubt the model architecture if the greatest predicted probability is arbitrarily close to the second greatest probability. An intuitive score that can capture this sense of certainty and doubt about our predictions is desirable.


\section*{Defining Certainty and Doubt}
Working in the context of (multi-class) classification decision problems on $N$ distinct labels, we will assume that we have a learning machine $f(\vect{\omega},\vect{x})$ that on input $\vect{x}\in \R^{Input}$ with weights $\vect{\omega}$, outputs a probability vector $\vect{p}\in\Delta^N$ after applying a differentiable function to normalize a vector $\vect{y}\in\R^N$, such as \textsc{SoftMax}, denoted by $\sigma$.

Let $j=\arg\max_i y_i$ and $\hat{p}=\max_i \pi_i(\vect{p})$, where $\pi_i$ are the canonical projection maps onto the $i^{th}$ component.

We define \textit{certainty} and \textit{doubt} as reciprocals, the geometric intuition behind this will be described in the next section. We define the \textit{pairwise certainty} and \textit{pairwise doubt} by
\begin{equation}\tag{Pairwise Certainty}\chi_{i}=\left\{\begin{matrix}1&i=j\\\pi_j(\vect{p})-\pi_i(\vect{p})\equiv(\pi_j-\pi_i)(\vect{p})&i\ne j\end{matrix}\right.\label{eqn:pwcertainty}\end{equation}
and
\begin{equation}\tag{Pairwise Doubt}\delta_{i}=\left\{\begin{matrix}0&i=j\\\cfrac{1}{\pi_j(\vect{p})-\pi_i(\vect{p})}&i\ne j\end{matrix}\right.\label{eqn:pwdoubt}\end{equation}
The intuitive motivation for these definitions is that if $f$ gives two components the same maximum score, then the model is `maximally' uncertain between the two labels, whence the certainty is 0 and doubt is infinite, while there should be no doubt when comparing one label score with itself; in order to reasonably doubt, an alternative must be possible. 
Our pairwise definitions can be used to define vector-valued maps from the probability simplex $\Delta^N$ to $\R^N$ and $\hat{\R}^N:= \R^{N}\cup\{\infty\}$ respectively via $\vect{\chi}=\chi_1\times\cdots\times \chi_N$ and $\vect{\delta}=\delta_1\times\cdots\times \delta_N$ for certainty and doubt, respectively. Moreover, the non-diagonal components of certainty and doubt are distributed between $[0,\hat{p}]$ and $[\hat{p}^{-1},\infty]$.

Further, for the ease of graphing the distribution of doubt scores, we may also consider either using $-\log$ post-composed with the pairwise certainty scores or the difference between the log-probabilities as a related notion of doubt. This latter definition can be straight forwardly computed without using log-softmax by computing $\hat{y}\mathbf{1}-\vect{y}$ before applying softmax, although care will need to be taken with respect to the index of maximal value. We define certainty and doubt in this setting as \textit{raw certainty} and \textit{raw doubt}. Formally:
\begin{equation}\tag{Pairwise Raw Certainty}\xi_i=\left\{\begin{matrix}1&i=j\\\pi_j(\vect{y})-\pi_i(\vect{y})\equiv (\pi_j-\pi_i)(\vect{y})&i\ne j\end{matrix}\right.\label{eqn:pwrc}\end{equation}
\begin{equation}\tag{Pairwise Raw Doubt}\rho_i=\left\{\begin{matrix}0&i=j\\\cfrac{1}{\pi_j(\vect{y})-\pi_i(\vect{y})}&i\ne j\end{matrix}\right.\label{eqn:pwrdoubt}\end{equation}

Finally, although it is more computationally expensive, we can also encode certainty across the assigned probability vector $\vect{p}$ with the following skew-symmetric matrix:
\[\vect{C}^o(\vect{p}):=\vect{p}\vect{1}^T-\vect{1}\vect{p}^T\]
and set certainty as
\[\vect{C}(\vect{p})=\vect{I}+\vect{C}^o(\vect{p}).\]

Since $\vect{C}(\vect{p})$ will always be invertible, we may naively suppose that $\vect{C}^{-1}$ is a good candidate for expressing doubt. However, consider the following binary classification vector $\vect{p}=\left[\begin{matrix}.5\\.5\end{matrix}\right]$. Then $\vect{C}=\vect{I}$ and $\vect{C}^{-1}=\vect{I}$, while intuitively we would be maximally uncertain between the two labels as the probability vector indicates they have equal likelihood. Instead, we may find doubt may be better described as
\[\vect{D}(\vect{p})=\mbox{Inv}(\vect{C}(\vect{p}))-\vect{I}\]
where $\mbox{Inv}$ is the element-wise inverse function. 

We propose developing scores from $\vect{C}(\vect{p})$ and $\vect{C}^o(\vect{p})$ for the purposes of quantifying uncertainty between different neural network architectures. For the ease of graphing the distribution of doubt and for studying the statistics of the doubt scores between models, we suggest setting our minimum certainty/maximum doubt score to be set to \[\min_i \pi_i(\vect{x}):\max\limits_{\vect{x}\in \mbox{Row}(\vect{C}(\vect{p}))}\|\vect{x}\|_1\hspace{1cm} \mbox{ or equivalently } \hspace{1cm} = \|\vect{D}(\vect{p})\|_\infty^{\prime},\] where $\|\vect{D}\|_\infty^{\prime}$ is the extended Chebyshev metric adding a point at infinity, with the goal of maximizing minimum certainty, or equivalently, minimizing maximum doubt. 


\section*{Geometric Interpretation}
\subsection*{Projective Geometry}
A \textit{projective space} over a field $K$ is a set of one-dimensional subspaces of the vector space $K^{n+1}$, which we will denote by $\mathbb{P}^n(K)$. This can be equivalently understood as the quotient of $K^{n+1}\backslash\{\vect{O}\}$ by the action of $K^\times$ acting by scalar multiplication, i.e. $\mathbb{P}^n(K)$ can be identified as a set of equivalence classes $[\vect{p}]$ such that any two non-zero points $\vect{p},\vect{q}$ in $\mathbb{P}^n(K)$ belong to the same equivalence class if $\vect{p}=c\vect{q}$ for some $c\in K^\times$. Points in projective vector spaces are usually written as \textit{homogeneous vectors} $[Z_0:\cdots:Z_N]$.

We are primarily interested in the geometry of \textit{real projective} spaces, which are typically denoted by $\R\mathbb{P}^n$. Real projective spaces are well-studied examples of real manifolds, with local coordinates $U_k$ of $\R\mathbb{P}^n$ given by
\[U_0=\{[1:x_1:x_2:\cdots:x_N]\mid x_1,x_2,\hdots,x_N\in\R\}\]
\[U_1=\{[x_0:1:x_2:\cdots:x_N]\mid x_1,x_2,\hdots,x_N\in\R\}\]
\[\vdots\]
\[U_N=\{[x_0:x_1:x_2:\cdots:1]\mid x_1,x_2,\hdots,x_N\in\R\}.\]
It is readily seen that each $U_i$ is analytically isomorphic to $\R^N$. Moreover,
\[\R\mathbb{P}^N\cong \R^N\cup \R^{N-1}\cup\cdots\cup \R\cup\R^0\]
which follows from the identification of
\[\R\mathbb{P}^N=U_0\cup\{[0:x_1:\cdots:x_N]\mid [x_1:\cdots:x_N]\in \R\mathbb{P}^{N-1}\}\cong \R^N\cup \R\mathbb{P}^{N-1}.\]

In particular, $\{[0:0:\cdots:0:1]\}\cong \R^0$ is referred to as \textit{the point at infinity}. 

Since spheres $S^N\cong \R\mathbb{P}^N$ by quotienting out by the antipodal points, and the standard Riemannian metric on $S^{N}$ is invariant under the action of the group $\mathbb{Z}/(2)$, we can push down the metric on $S^N$ to $\R\mathbb{P}^N$. Explicitly, the metric on $\R\mathbb{P}^N$ is defined by
\[\langle v,w\rangle_{[p]}=\left\langle(dq|_p)^{-1}(v),(dq|_p)^{-1}(w)\right\rangle_p,\]
with $q:S^N\to S^N/\sim_\alpha\cong \R\mathbb{P}^N$ the quotient map of $S^N$ by the antipodal equivalence relation $\sim_\alpha$ and $[p]=\{\pm p\}$ for $p\in S^N$. Further, this is well defined since the antipodal map $\alpha:S^N\to S^N$ is an isometry, so $q=q\circ \alpha$, and $d\alpha_p$ will also be an isometry.

\subsection*{The Geometry of Doubt}

Certainty and doubt can be naturally viewed using projective geometry.

Consider the case where we have $N$ labels, and $\Delta^N$ is the probability simplex, e.g. \[\Delta^N=\{\vect{p}\in \R^N\mid \mathbf{1}_N\cdot \vect{p} =1\land \bigwedge\limits_{1\le i\le N}\pi_i(\vect{p})\ge 0\},\]
where $\pi_i\colon \R^N\to\R$ is the canonical $i^{th}$-coordinate projection function. 

Further, we may always permute the indices as necessary so that $\mathbf{p}$ is such that $\pi_i(\vect{p})\ge\pi_{i+1}(\vect{p})$ for $i\in[N]_+$. Denote $\Delta^+_\ge$ to be this ordered simplex. 

 We let $F\colon\Delta^N_{\ge}\to\prod\limits_{i\in[N]_+}\R\mathbb{P}^1$ be given coordinate wise by 
 \[f_i(\vect{p})=\left\{\begin{matrix}[1:0]& i=1\\ [\pi_1(\vect{p})-\pi_i(\vect{p}):1] & i\ne 1\end{matrix}\right.\] 
 
 Since the first coordinate we're mapping is the distance between $p_1$ and $p_i$, this can be understood as expressing our `certainty' that the first label is correct over label $i$, while the second coordinate can be thought of as expressing our doubt. In particular, we have absolute certainty when comparing one probability with itself, as there is no point of comparison. Morevoer, since $[p_1-p_i:1]$ is a point in an equivalence class, we see that $[p_1-p_i:1]=[1:\frac{1}{p_1-p_i}]$ whenever $p_1>p_i$, so that pairwise doubt and certainty identify the same point on the real projective line, while whenever $p_1=p_i$, $f(p_i)=[0:1]$, i.e. we have a point at infinity, indicating we have infinite doubt and zero certainty that our top choice is in fact the correct one. 
 
Further, because the image of $\Delta^N_\ge$ in $ (\R\mathbb{P}^1)^N$ under $F$ is identical to $\{[1:0]\}\times (\R\mathbb{P}^1\backslash\{[1:0]\})^{N-1}$, the image will be isomorphic with $(\R\mathbb{P}^1\backslash\{[1:0]\})^{N-1}$. Thus we can consider $F:\Delta^N_\ge\to \prod\limits_{i=2}^N\R\mathbb{P}^1$ or alternately, $F^\prime :\Delta^N_\ge \to \prod\limits_{i=2}^{N}S^1$ instead. 

\section*{Proposed Cost Function With Certainty and Doubt Scoring}

Taking the projective view, doubt-minimization is a program that aims to avoid estimated probability assignments whose Segre embedding is a degenerate point lying in a real-projective space. In particular, we wish to avoid $\prod\limits_{i=2}^N(p_1-p_i)=0$, which indicates that we have no confidence between our assigned label and another label choice. On the other hand, there is no intrinsic reason that we should prefer a non-uniform distribution on the other assigned probabilities below the maximum estimated probability.

When computing a cost function that invokes the doubt, or certainty of a score, we wish to have some appropriate smooth function. We propose using the stereographic projection diffeomorphism $F:S^1\to\R\mathbb{P}$ given by sending $\theta\mapsto\left\{ \begin{matrix} [1-\sin\theta : \cos \theta] & \theta\ne \frac{\pi}{2} \\ [\cos\theta:1+\sin\theta]& \theta\ne -\frac{\pi}{2}\end{matrix}\right.$, with inverse $G:\R\mathbb{P}\to S^1$ given by $$[a:b]\mapsto\left\{ \begin{matrix} \arcsin\left( \cfrac{b^2-a^2}{a^2+b^2}\right) & a\ne 0 \\ \frac{\pi}{2}& [a:b]=[0:1]\end{matrix}\right. .$$ 
Two natural candidates for cost per sample $k$ are the doubt-cost $\theta_{cost}(\vect{p}^{(k)})=\arcsin \left(\cfrac{1-\prod\limits_{i\ne j}\chi_i^2}{1+\prod\limits_{i\ne j}\chi_i^2}\right)$, and the raw-doubt cost $\theta_{cost}(\vect{y}^{(k)})=\arcsin \left(\cfrac{1-\prod\limits_{i\ne j}\xi_i^2}{1+\prod\limits_{i\ne j}\xi_i^2}\right)$. The cost functions incorporating doubt should be reflective of the underlying function/phenomenon that $f$ is trying to approximate, and penalize doubt accordingly.



\section*{Acknowledgements}
This work is supported in part by the U.S. Army Combat Capabilities Development Command (DEVCOM) Army Research Laboratory under Support Agreement No. USMA 21050, as well as the Defense Advanced Research Projects Agency (DARPA) under Support Agreement No. USMA 23004. The views expressed in this paper are those of the authors and do not reflect the official policy or position of the United States Military Academy, the United States Army, the Department of Defense, or the United States Government.

\printbibliography

\end{document}